%% file: _main.tex
\documentclass[10pt,twocolumn,letterpaper]{article}

\usepackage{cvpr}

\usepackage{ulem}
\usepackage{comment}
\usepackage{tabularx}
\usepackage{booktabs}
\usepackage{xurl}
\usepackage[accsupp]{axessibility}
\usepackage[T1]{fontenc}

\input{preamble}
\definecolor{cvprblue}{rgb}{0.21,0.49,0.74}
\usepackage[pagebackref,breaklinks,colorlinks,allcolors=cvprblue]{hyperref}

\newcommand{\dataset}{CSA-Graphs}

\def\confName{CVPR}
\def\confYear{2026}

\title{\dataset: A Privacy-Preserving Structural Dataset\\for Child Sexual Abuse Research}

\author{Carlos Caetano$^{1}$\thanks{Corresponding authors: \{caetanoc, avilas\}@unicamp.br} \and Camila Laranjeira$^{2}$ \and Clara Ernesto$^{3}$ \and Artur Barros$^{1}$ \and João Macedo$^{4, 5}$ \and Leo S. F. Ribeiro$^{3}$ \and Jefersson  A. dos Santos$^{6}$ \and Sandra Avila$^{1 *}$
\and \\
\small$^{1}$Universidade Estadual de Campinas (UNICAMP), Brazil \\
\small$^{2}$Instituto Federal de Educação, Ciência e Tecnologia de Minas Gerais (IFMG), Brazil \\
\small$^{3}$Universidade de São Paulo (USP), Brazil \quad
\small$^{4}$Universidade Federal de Minas Gerais (UFMG), Brazil \\
\small$^{5}$Polícia Federal (PF), Brazil \quad
\small$^{6}$University of Sheffield, England, United Kingdom
}

\begin{document}
\maketitle
\input{0_abstract}    
\input{1_introduction}
\input{2_related_works}
\input{3_RCPD_dataset}
\input{4_representations}
\input{5_baseline_experiments}
\input{6_conclusion}
\input{7_ethical_legal_considerations}
\input{9_acknowledgments}
{
    \small
    \bibliographystyle{ieeenat_fullname}
    \bibliography{references}
}

\input{8_supplementary}

\end{document}

%% file: 0_abstract.tex
\begin{abstract}

Child Sexual Abuse Imagery (CSAI) classification is an important yet challenging problem for computer vision research due to the strict legal and ethical restrictions that prevent the public sharing of CSAI datasets. This limitation hinders reproducibility and slows progress in developing automated methods. In this work, we introduce \dataset, a privacy-preserving structural dataset. Instead of releasing the original images, we provide structural representations that remove explicit visual content while preserving contextual information. \text{\dataset} includes two complementary graph-based modalities: scene graphs describing object relationships and skeleton graphs encoding human pose. Experiments show that both representations retain useful information for classifying CSAI, and that combining them further improves performance. This dataset enables broader research on computer vision methods for child safety while respecting legal and ethical constraints.

\end{abstract}

%% file: 1_introduction.tex
\section{Introduction}
\label{sec:introduction}

Child Sexual Abuse Imagery (CSAI), also referred to as Child Sexual Abuse Material (CSAM)~\cite{Greijer:CSAM_Terminology:2016}, represents one of the most severe forms of harmful content circulating in digital environments. The proliferation of online platforms and digital storage devices has significantly increased the scale at which such material can be produced, distributed, and shared, creating major challenges for law enforcement agencies (LEAs) and online service providers. Identifying and removing CSAI from digital systems is therefore a critical task for protecting victims and preventing further dissemination. In recent years, computer vision and machine learning techniques have been increasingly explored to assist investigators and moderation systems in identifying abusive content at scale~\cite{Lee2020, gangwar2021attm, Laranjeira:FAccT:2022, rondeau2022deep, Coelho:SIBGRAPI:2024, Valois:FSI:2025, Barros:BMVCW:2025, Coelho:FAccT:2025}. However, developing and evaluating such methods presents unique challenges due to the legal and ethical restrictions surrounding access to CSA~data\footnote{Producing, possessing, or distributing CSAM is a criminal offense in many jurisdictions and is subject to strict legal controls~\cite{DepJusticeUS:CSAM:2023, criminal_code_australia_1995, lei_11829_2008, EU2011Directive93}.}.

Law-enforcement pipelines for investigating seized devices, as well as content moderation systems operated by social media platforms and service providers, have historically benefited from databases containing unique fingerprints (hash values) of CSAI previously reported and confirmed by human experts~\cite{Lee2020}. Through hash-matching techniques such as PhotoDNA~\cite{PhotoDNA_2011}, one can quickly and effectively identify repeated instances of abusive content. 

Because these hashes do not reconstruct or reveal the underlying visual content, such databases can be shared among authorized institutions under less restrictive conditions than the raw images themselves, enabling collaboration while reducing the risks associated with distributing sensitive material. The National Center for Missing and Exploited Children (NCMEC) reported that, in 2024, nearly 10~million hash values were shared with dozens of technology companies that wish to report and remove CSAI shared through their systems \cite{nmec_2024}. 

On the other hand, the literature on identifying novel CSAIs lacks a similar advantage. Datasets that support the development of CSAI classification tools are often the product of local partnerships with LEAs, which provide indirect access to seized material from investigations for training or evaluating techniques~\cite{macedo2025child}. Unlike hash databases, which are unique sources of data shared by dozens of initiatives, datasets containing raw CSAI images are rarely reused across studies, limiting direct comparability.

This raises the question of how to conduct computer vision research on illegal imagery, allowing scientific scrutiny while respecting legal and ethical constraints. A common strategy is to work with sub-tasks within the target illegal domain, such as combining age estimation models with some form of indecency classification~\cite{sae2014towards, Macedo:RCPD:2018, gangwar2021attm, rondeau2022deep}, or exploring contextual cues from the scene~\cite{Coelho:SIBGRAPI:2024, Valois:FSI:2025, Barros:BMVCW:2025}. These related tasks were later formalized under the name of \textit{Proxy Tasks} as a path for risk minimization for working with illegal imagery~\cite{Coelho:FAccT:2025}. However, experimenting on images from the target domain is paramount for evaluation, and therefore, strict data access restrictions remain an issue.

\begin{figure}[t]
   \centering
    \subfloat[]{
    \includegraphics[width=0.32\linewidth]{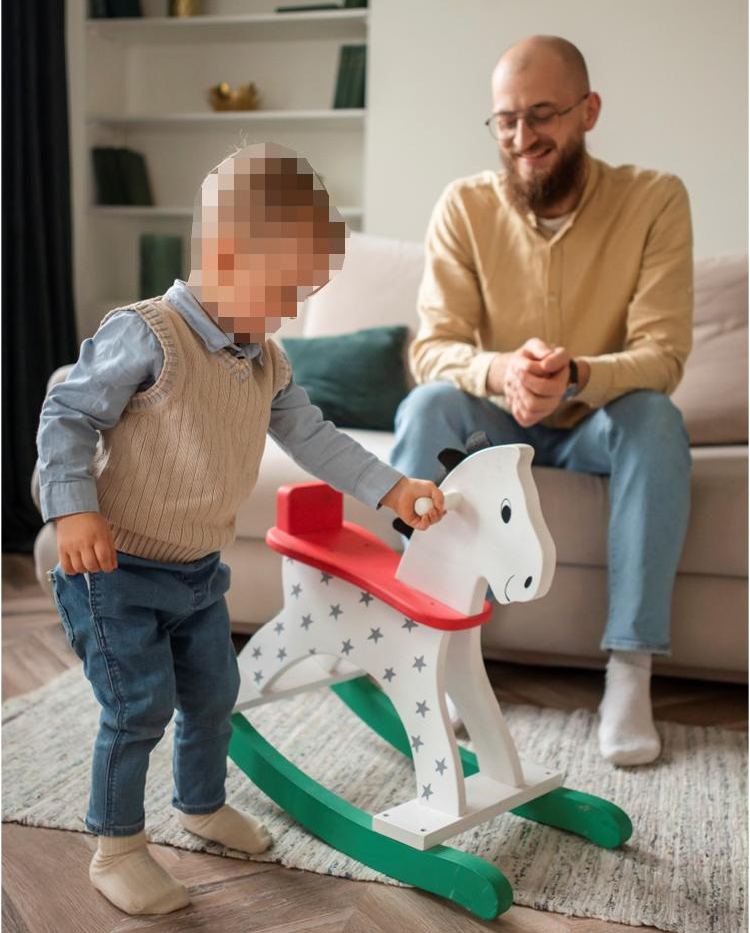}\label{fig:img_skl_sg_a}}\hspace{0.01cm} 
    \subfloat[]{\includegraphics[width=0.32\linewidth]{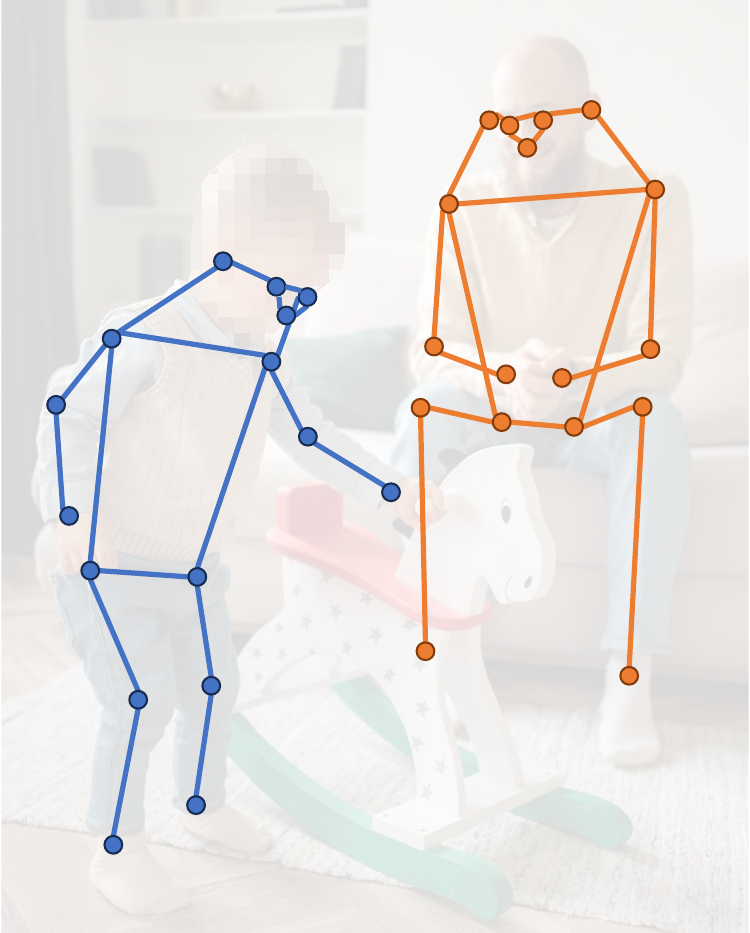}\label{fig:img_skl_sg_b}}\hspace{0.01cm} 
    \subfloat[]{\includegraphics[width=0.32\linewidth]{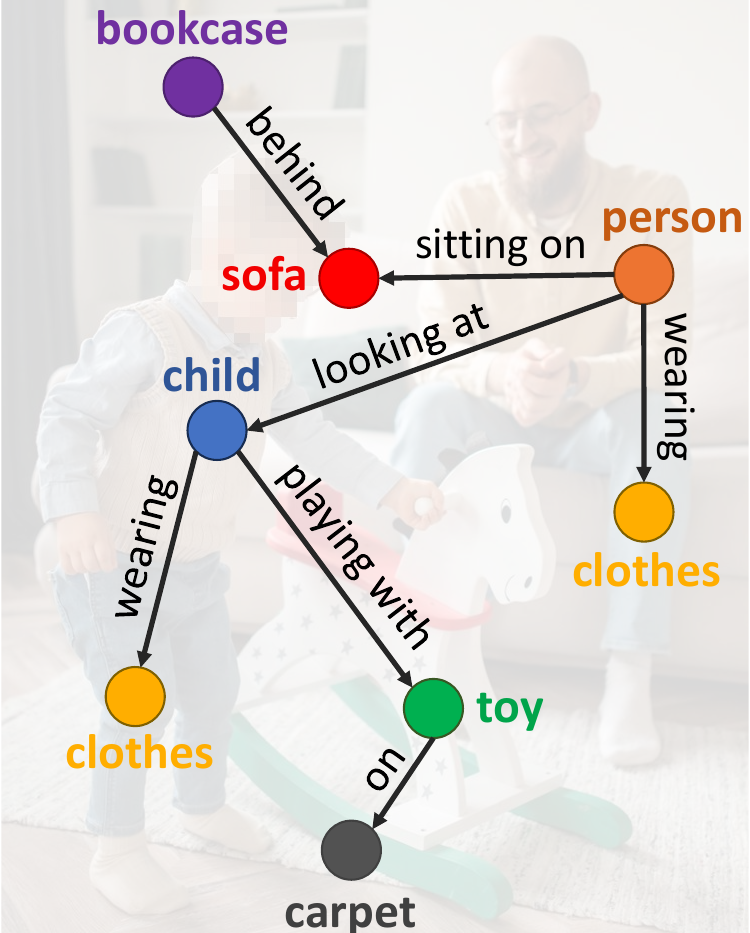}\label{fig:img_skl_sg_c}}
   \caption{Example of the graph-based representations used in this work. (a) Original image. (b) Skeleton pose graph obtained through human pose estimation. (c) Scene graph describing objects and their relationships in the scene. The graphs are overlaid on the original image for visualization purposes only.
   }
   \label{fig:img_skl_sg}
   \vspace{-10pt}
\end{figure}

In this work, we address this challenge by introducing \dataset, a privacy-preserving dataset derived from the Region-Based Annotated Child Pornography Dataset (RCPD)~\cite{Macedo:RCPD:2018}, in which the original images are replaced by structural representations that capture relevant contextual information without exposing visual content. Specifically, we generate two complementary graph-based representations, as illustrated in \text{\autoref{fig:img_skl_sg}}: (i)~scene~graphs~\cite{Li:sg_survey:Neuricomputing:2024}, which describe objects and their relationships within a scene; and (ii)~human skeleton~pose~graphs~\cite{Zheng:skl_survey:CompSurv:2023}, which represent body pose through anatomical keypoints. These structured representations have been widely used in computer vision tasks involving scene understanding~\cite{Nie:ECCV:2022, Khandelwal:scene_underst:Neurips:2024}, human interaction analysis~\cite{He:ExploitingSG:ICCV:2021, Li:CVPR:2024}, activity recognition~\cite{Rodin:Action_SG:CVPR:2024, Aich:ICCV:2023}, and sensitive media recognition~\cite{Antonia:WCAID:2024, Barros:BMVCW:2025}, where relational and pose information provide important contextual cues. By abstracting images into relational and pose-based structures, these representations retain contextual and interaction cues that are useful for studying CSAI-related patterns while preventing direct access to the underlying imagery. 

Our main contributions are summarized as follows:
\begin{itemize}

    \item \textbf{\dataset~dataset release.} We introduce a publicly accessible dataset comprising scene graphs~\cite{Caetano:SIBGRAPI:2024} and human-skeleton pose~\cite{Cao:CVPR:2017} representations derived from the RCPD dataset~\cite{Macedo:RCPD:2018}, a law-enforcement resource maintained by the Brazilian Federal Police that contains both CSAI and non-sensitive imagery. By providing two complementary modalities, the dataset enables investigation of contextual and relational patterns of CSAI, without exposing harmful content.

    \item \textbf{First exploration of skeleton-based representations for CSAI classification.} To our knowledge, we present the first investigation of human pose graphs for CSAI classification, showing that pose-based structural features encode meaningful cues related to abusive scenarios.

    \item \textbf{Experimental validation and modality complementarity.} By employing a Graph Attention Network (GAT) baseline, we show that both representations retain useful information for CSAI classification and that combining scene graphs and skeleton poses improves classification performance. The code is available\footnote{\url{https://github.com/araceli-project/CSA-Graphs}}.

    \item \textbf{Enabling broader research on child safety in computer vision.} By releasing these derived representations, we lower the barrier for researchers to develop and evaluate models for sensitive content classification, fostering reproducibility and encouraging further work on child safety applications in computer vision.

\end{itemize}

%% file: 2_related_works.tex
\section{Related Work}
\label{sec:related_works}

A recent survey on CSAI datasets~\cite{macedo2025child} reviewed computer vision techniques leveraging real or simulated imagery for tasks such as victim identification, statistical analysis, and, overwhelmingly, CSAI classification. A key criticism by the authors is that CSAI datasets are often a product of local partnerships with LEAs, providing indirect access to seized material from investigations, with little to no intervention in data preparation and quality assessment. Therefore, datasets lack guarantees of task complexity, attribute diversity, and, most importantly, they hinder scientific scrutiny.

\begin{table*}[!ht]
    \footnotesize
    \centering
    \caption{Datasets containing real CSAI, summarized by year of introduction, works citing them, LEA origin, size, and label scheme. 
    Information compiled from \citet{macedo2025child}. LEA acronyms: BR-FP (Brazilian Federal Police), EU (European Union), AU-FP (Australian Federal Police), ES-NC (Spanish National Cybersecurity Institute), MT-PF (Malta Police Force), US (United States).}
    \begin{tabularx}{\textwidth}{lc>{\raggedright}p{1.4cm}p{0.9cm}X }
      \hline
      Dataset & Year & Works & LEA & Description \\
      \hline
      PE1-A & 2010 & \cite{de2010nudetective} & BR-FP & 80GB seized hard disk including novel CSAI. \\
      ULS-A & 2011 & \cite{ulges2011automatic} & EU &  36,198 images (20,000 CSAI) classified as CSAI, porn, or world.\\
      SCH-A & 2014 & \cite{schulze2014automatic, peersman2016icop} & EU & 60,000 images (20,000 CSAI) classified as CSAI, adult, or world. \\
      RealCSA & 2017 & \cite{gangwar2017pornography} & ES-NC & 5,000 images (2,500 CSAI) of juveniles.\\
      VTR-A & 2018 & \cite{vitorino2018leveraging} & BR-FP & Seized hard disk comprising 58,974 images with 33,723 CSAI. \\
      RCPD & 2018  & \cite{Macedo:RCPD:2018, Laranjeira:FAccT:2022, Coelho:SIBGRAPI:2024, Valois:FSI:2025, Barros:BMVCW:2025, Coelho:FAccT:2025, Laranjeira:2026humancentricperceptionchildsexual} & BR-FP & 2,138 images (836 CSAI) with region-based labels for age, gender, and nudity levels. \\
      CEM Corpora & 2018 & \cite{dalins2018laying} & AU-FP & Over 100,000 images from 13 cases, with CSAI severity labels. \\
      TorCrawl & 2018 & \cite{dalins2018laying} & AU-FP & Images extracted from 232,792 Tor pages, including CSAI. \\
      Test-Set-CSA & 2021 & \cite{gangwar2021attm} & ES-NC & 2,200,379 images (5,000 CSAI) classified as CSAI, porn, juvenile, and general. \\ 
      CSA & 2021 & \cite{tabone2021pornographic} & MT-PF  & Unlabelled set of 384 CSA images.  \\
      RDU-A & 2022 & \cite{rondeau2022deep} & US & Over 80,000 images, not guaranteed to contain children and nudity.  \\
      ORO-A & 2024 & \cite{OronowiczJakowiak2024} & --  & 60,000 images (15,000 CSAI) labeled by forensic experts as CSAI, pornography, people, or other.\\
      CSAI & 2025 & \cite{Coelho:FAccT:2025} & BR-FP & 4,592 images classified as CSAI, suspected CSAI, pornography, people, drawing, other.  \\
      \hline
    \end{tabularx}
    \label{tab:csai-datasets}
    \vspace{-10pt}
\end{table*}

\autoref{tab:csai-datasets}, derived from the systematic review in \citet{macedo2025child}, lists all datasets containing real images of child sexual abuse. To our knowledge, all such datasets comprise raw images, with no instances of privacy-preserving derived datasets for the CSAI classification task.

Some datasets in \autoref{tab:csai-datasets} are structured to provide a challenging classification scenario, supplementing CSAI acquired from LEA partners with images from additional data sources. For instance, \citet{ulges2011automatic} gathered adult pornography and several non-sensitive images crawled from public websites to represent the classes porn and world. It allowed for a variety of test cases, which helped establish adult pornography as the most challenging category to distinguish from CSAI. A similar protocol is followed by other works~\cite{schulze2014automatic, peersman2016icop, Macedo:RCPD:2018, gangwar2021attm, OronowiczJakowiak2024}, all of which highlight that CSA samples were handled by authorized personnel, and that other researchers did not access raw images.

While most CSAI datasets are superficially presented, with key information unknown even to the researchers, important exceptions stand out. \citet{dalins2018laying} leverage a dataset labeled by forensic experts with high granularity and diversity of labels, according to the Child Exploitation Tracking System (CETS), a 10-point scale of CSAI severity, classified by the activities, posing, objects, and context depicted. Similarly, \citet{OronowiczJakowiak2024} invited a forensic psychologist, sexologist, and anthropologist to label images using the anthropometric SMR scale for patterns of pubertal changes and the COPINE scale for CSAI severity. To provide explainability, the authors evaluated the relation between class activation maps and anatomical body parts deemed important. Lastly, RCPD~\cite{Macedo:RCPD:2018} provides region-based labels for demographic attributes and nudity levels, enabling various tasks and offering the additional advantage of serving as a publicly available benchmark. The authors released the LEA’s contact and technical details of the dataset, enabling the submission of ready-to-run tools. It currently amounts to seven works that have been experimented on RCPD~\cite{Macedo:RCPD:2018, Laranjeira:FAccT:2022, Coelho:SIBGRAPI:2024, Barros:BMVCW:2025, Valois:FSI:2025, Coelho:FAccT:2025, Laranjeira:2026humancentricperceptionchildsexual}.

Previous efforts to improve CSAI datasets have focused on label characteristics, such as the granularity of categories and the diversity of annotated attributes. However, working with raw images still requires indirect access through LEA partners. As a consequence, the majority of works in  \text{\autoref{tab:csai-datasets}} are used in a single study, lacking comparability and severely limiting their contribution to quality assessment.

Due to access restrictions, CSAI datasets are often leveraged as the final experimentation stage, evaluating models produced with the support of legally accessible datasets~\cite{gangwar2017pornography, Macedo:RCPD:2018, gangwar2021attm, tabone2021pornographic, Valois:FSI:2025}, sometimes fine-tuning on the target domain~\cite{Coelho:SIBGRAPI:2024, Coelho:FAccT:2025, Barros:BMVCW:2025}. For instance, training separate models for age estimation and pornography classification~\cite{Macedo:RCPD:2018, gangwar2021attm} enables the use of publicly available datasets during training, leaving the illegal imagery only for testing. A recent research trend also explores the importance of contextual cues for CSAI classification~\cite{Laranjeira:FAccT:2022, Valois:FSI:2025, Barros:BMVCW:2025, Coelho:FAccT:2025}, pursuing scene classification as the sub-task of interest.

\citet{Coelho:FAccT:2025} formalized the aforementioned approach as a risk minimization technique leveraging \textit{proxy tasks}, i.e., sub-tasks related to the illegal domain that allow training with legal datasets. Although proxy tasks are a valid strategy, training on the target domain has many benefits. Notably, \citet{vitorino2018leveraging} experiment with models trained for pornography classification and fine-tuned on CSAI. While the widely used Yahoo NSFW model~\cite{mahadeokar2016open} achieves $77.5\%$ accuracy on the CSAI dataset, a model trained directly on the target domain reaches up to $85.5\%$. 

In summary, CSAI classification research faces severe challenges in data access, leading to lack of information on dataset quality, making it unfeasible to compare studies, and pushing research towards using images from related domains. Given that raw images cannot be made public, methods grounded in derived representations of target-domain data constitute a promising avenue for advancing the field.

%% file: 3_RCPD_dataset.tex
\section{RCPD and Legal Constraints}
\label{sec:rcpd}

The Region-Based Annotated Child Pornography Dataset~(RCPD)~\cite{Macedo:RCPD:2018} was developed through a collaboration between academic researchers and the Brazilian Federal Police to support research on automated classification of CSAI. Due to the illicit nature of the dataset, the images cannot be publicly distributed or illustrated in scientific publications. The dataset was originally designed to serve as a benchmark for evaluating computational methods for classifying CSAI in digital~media.

RCPD consists of 2,138 images across different types of content, including CSAI, adult pornography, and non-sensitive images. Of these, 1,630 depict individuals, while the remaining images consist of scenes without people, enabling false positive evaluation. The dataset does not contain classification labels for CSAI, instead, the images were annotated by computer forensic specialists with region-based labels describing the presence of individuals and specific body parts, along with semantic labels related to perceived demographic attributes and visual characteristics.

Specifically, these annotations comprise bounding boxes over persons, faces, and relevant unclothed regions,  resulting in thousands of annotated objects across the dataset. Each individual is further labeled with an estimated age range, perceived gender, and degree of nudity or sexual interaction. This combination enables multiple evaluation tasks, such as identifying the presence of minors in images containing explicit content or distinguishing CSAI from other types of imagery. For instance, in \text{\citet{Macedo:RCPD:2018}}, CSAI classification labels are derived from images depicting nudity or sexual interactions and the presence of a child aged 13 or younger. This annotation has established RCPD as a benchmark for CSAI classification~\cite{Laranjeira:FAccT:2022, Coelho:SIBGRAPI:2024, Barros:BMVCW:2025, Valois:FSI:2025, Coelho:FAccT:2025, Laranjeira:2026humancentricperceptionchildsexual}.

\vspace{-3mm}

\paragraph{Access Restrictions.} The illegal nature of CSAI imposes strict restrictions on the storage and access of RCPD~\cite{DepJusticeUS:CSAM:2023, lei_11829_2008, EU2011Directive93}. Images are held exclusively by the Brazilian Federal Police, and academic researchers have no direct access to them at any point. As the dataset cannot be distributed, conventional download-and-experiment workflows are not possible, substantially limiting reproducibility and constraining real-world CSAI classification research.

\vspace{-3mm}

\paragraph{Secure Experimental Protocol.} Research using RCPD follows a controlled protocol developed in collaboration with law enforcement authorities, under which the dataset remains within LEA's secure infrastructure at all times. Researchers submit code or executable models, which authorized personnel execute on their behalf, returning only experimental results. While this procedure allows evaluation of models without exposing researchers to illegal material, it also introduces practical limitations, such as the difficulty of sharing datasets or reproducibility.

%% file: 4_representations.tex
\section{\dataset}
\label{sec:representations}

To address the scientific challenges related to strict constraints on CSAI data access, we introduce publicly shareable structural representations derived from RCPD that retain contextual and relational information without exposing sensitive content.
We generate two complementary representations: scene graphs and human skeleton graphs. Scene graphs describe objects and their relationships within a scene, while skeleton graphs encode human pose through body keypoints and their connections. As discussed in Section~\ref{sec:rcpd}, although RCPD contains 2,138 images, only 1,630 include individuals. We restrict the dataset to images with people. As a result, \dataset\ comprise representations extracted from 1,630 images: 837 CSAI and 793 non-CSAI. 

\subsection{Scene Graph Extraction}

Scene graphs provide a structured representation of visual scenes by modeling objects and their relationships as a graph~\cite{Caetano:SIBGRAPI:2024}. Although relational descriptions of images have been explored in earlier works, the concept of scene graphs was formally introduced by \citet{Johnson:CVPR:2015}. In this formalism, nodes denotes  objects in the image, while edges encode semantic relationships between pairs of objects, typically expressed as  $<$subject, predicate, object$>$ triplets. By explicitly capturing interactions between entities, scene graphs provide a compact abstraction of the scene that preserves contextual and relational information while removing the underlying pixel-level representation. \autoref{fig:img_skl_sg_c}~illustrates a scene graph extracted from \autoref{fig:img_skl_sg_a}.

Prior work explored approaches to analyze CSAM datasets without exposing the underlying images~\cite{Laranjeira:FAccT:2022}. Their study highlighted that identifying abusive content often depends not only on isolated visual cues such as nudity or age, but also on contextual factors including scene composition, surrounding objects, and environmental characteristics. These findings are consistent with reports from law enforcement investigations \cite{Kloess:ChallengesCSAM:2019,Kloess:ChallengesCSAM:2021}. Structured representations that encode objects and relationships, therefore, offer a promising alternative for data analysis whithout exposing the original imagery.

To generate scene graphs from the RCPD images, we employ the Pix2Grp framework proposed by \citet{Li:Pix2Grp:CVPR:2024}. Pix2Grp formulates scene graph generation as an image-to-sequence task using a vision-language model (VLM). Instead of relying on conventional detector-based pipelines, the framework directly generates relation triplets from the input image using a generative VLM.

As suggested by \citet{Li:Pix2Grp:CVPR:2024}, we adopt BLIP as the vision–language backbone, which combines a Vision Transformer (ViT-B/16) visual encoder with a BERT-based text decoder. We use the Pix2Grp model weights publicly released by the authors, pre-trained on the VG150~benchmark~\cite{Krishna:IJCV:2017} derived from the Visual Genome dataset.

The model outputs relation triplets composed of a subject, a predicate describing the relationship, and an object. These triplets are then used to construct the scene graph for each image. Because the model is trained on VG150, the predicted labels are restricted to the 150 object categories and 50 relationship predicates defined in that benchmark. This predefined vocabulary does not include object categories or relationship predicates directly associated with CSAI. Consequently, the generated graphs capture general contextual interactions between entities while avoiding explicit semantic descriptions of sensitive content.

By releasing structured abstractions instead of the original images, we enable the analysis of contextual and relational patterns while preserving the legal and ethical constraints associated with CSAI.

\subsubsection{Scene Graph Statistics}
\label{scene_graph_statistics}

\autoref{tab:sc_summary} summarizes the main statistics of the scene graph representations in CSA-Graphs. It contains 1,630~images represented as scene graphs, comprising 82,983 relational triplets. Among these, 40,624 triplets correspond to CSAI and 42,359 to non-CSAI. The scene graphs span 147 distinct object classes and 37 predicate classes derived from the VG150 vocabulary. On average, each image contains 50.91 triplets, with similar densities observed for both CSAI and non-CSAI subsets. Each image also contains an average of 26.73 unique object instances and about 10 unique predicates, indicating a relatively rich relational structure per scene. We computed all statistics~using a prediction score threshold of 0.50 for the generated~triplets.

\begin{figure}[h]
   \centering
   \subfloat[Top-10 object classes]{
       \includegraphics[width=0.9\linewidth, clip,trim={0.75cm 0.75cm 0.75cm 0.75cm}]{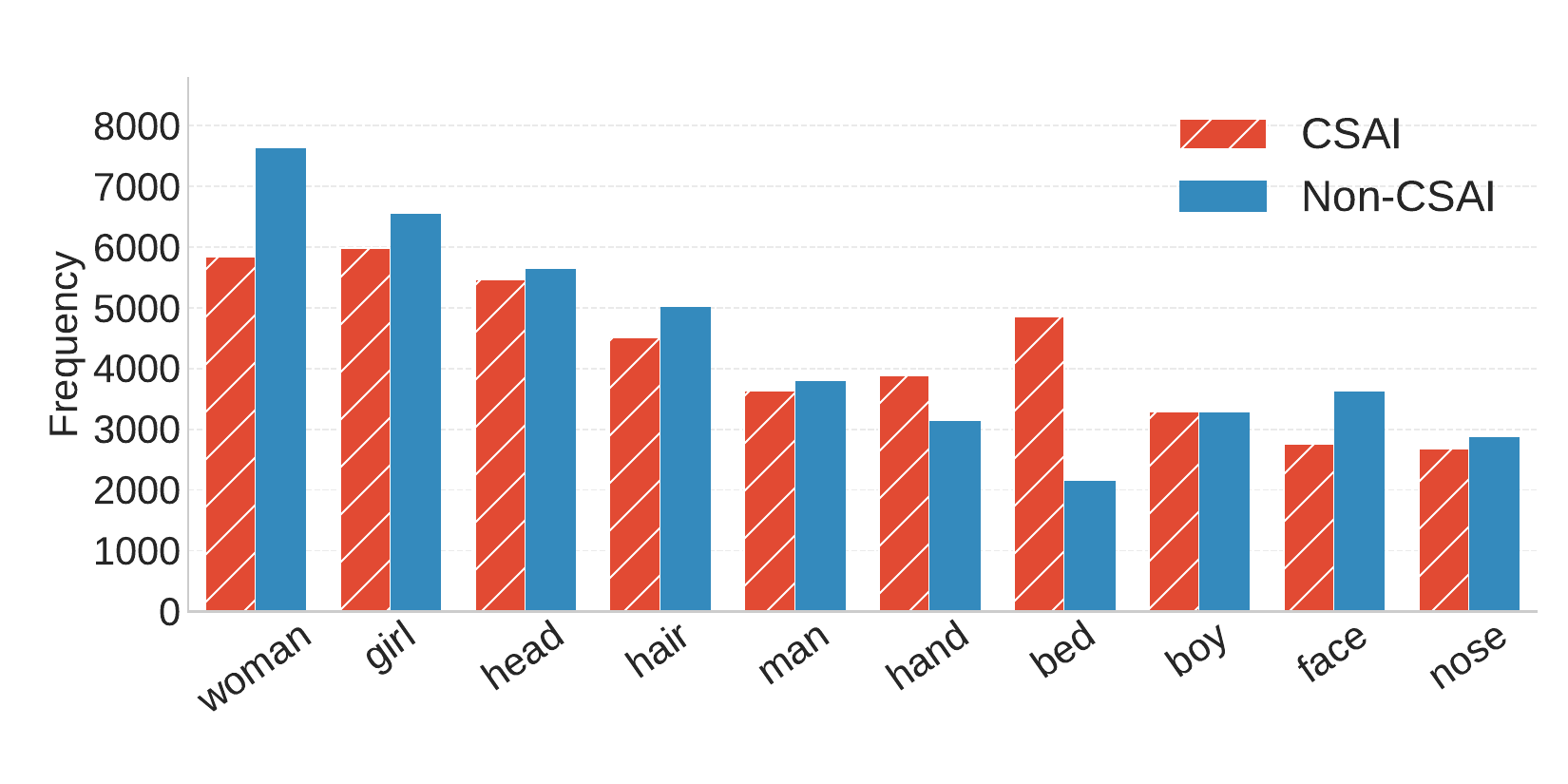}
       \label{fig:histograms_SG_a}
   }\\
   \subfloat[Top-10 predicates]{
       \includegraphics[width=0.9\linewidth, clip,trim={0.25cm 0.25cm 0.5cm 0.25cm}]{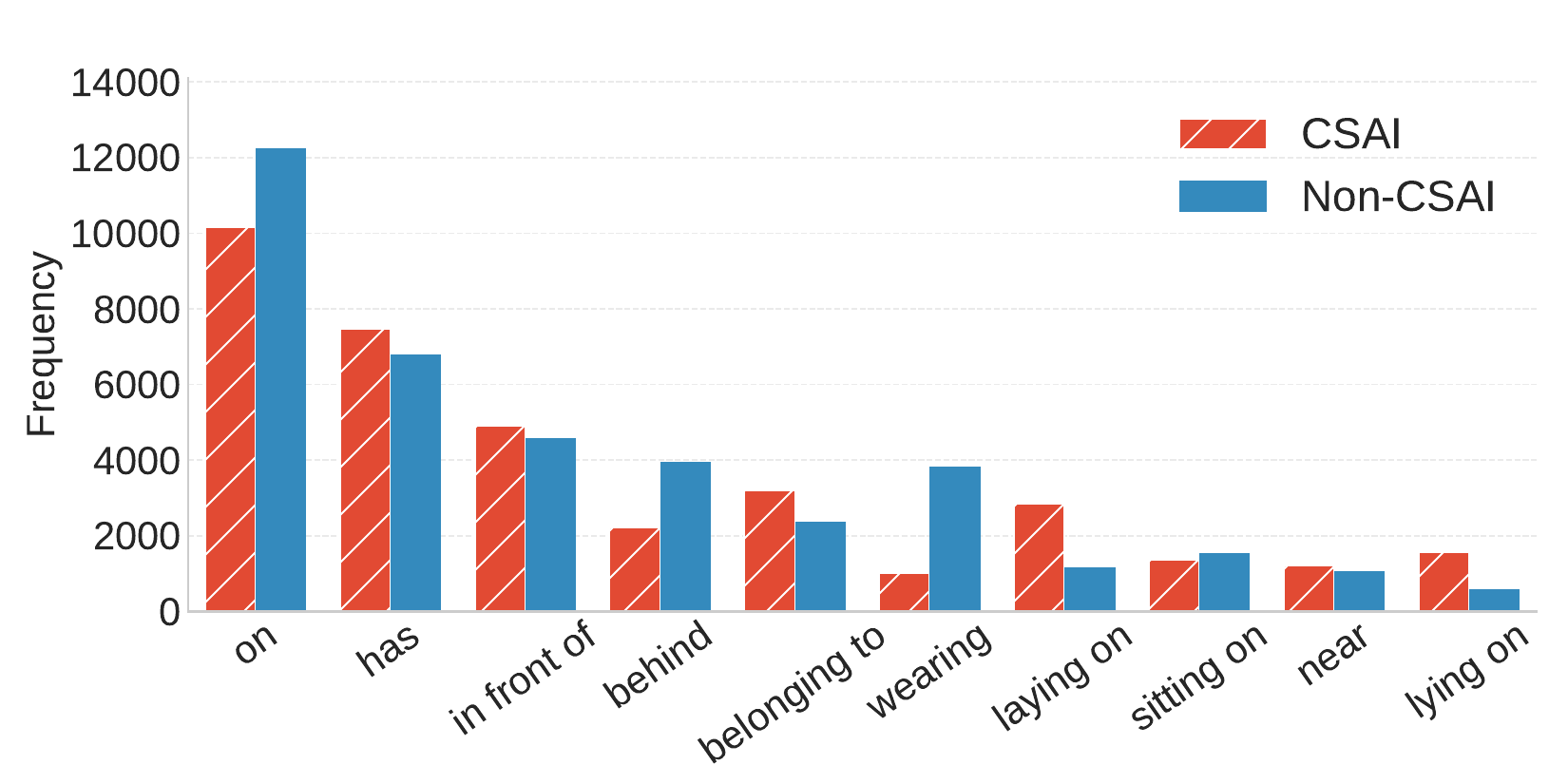}
       \label{fig:histograms_SG_b}
   }

   \caption{Most frequent scene graph elements in \dataset.}
   \label{fig:histograms_SG}
   \vspace{-10pt}
\end{figure}

\autoref{fig:histograms_SG} illustrates the distribution of the most frequent scene graph elements in \dataset. As shown in \text{\autoref{fig:histograms_SG_a}}, object categories related to children, such as \textit{girl} and \textit{boy}, appear among the most frequent detections, which is expected given the nature of the dataset. The distribution further highlights how these objects appear across both CSAI and non-CSAI subsets.

\autoref{fig:histograms_SG_b} presents the distribution of the most frequent relational predicates. In particular, the predicate \textit{wearing} appears considerably less frequently in CSAI images, which may reflect the lower presence of clothing in abusive scenarios. Predicates such as \textit{laying on} also appear among the most frequent relations, capturing pose configurations and interactions between individuals and objects.

\begin{table}[t]
    \footnotesize
    \centering
    \caption{Statistics of the scene graphs in \text{\dataset}.}
    \begin{tabular}{lccc}
        \hline
         & \textbf{All} & \textbf{CSAI} & \textbf{Non-CSAI} \\
        \hline
        Triplets & 82,983 & 40,624 & 42,359 \\
        Object classes & 147 & -- & -- \\
        Predicate classes & 37 & -- & -- \\
        Avg.~triplets / image & 50.91 & 48.54 & 53.42 \\
        Avg.~unique objects / image & 26.73 & 26.29 & 27.20 \\
        Avg.~unique predicates / image & 10.08 & 10.18 & 9.98 \\
        \hline
    \end{tabular}
    \label{tab:sc_summary}
    \vspace{-10pt}
\end{table}

\subsection{Skeleton Pose Extraction}

Human pose representations, or skeletons~\cite{Cao:CVPR:2017}, provide a structured description of the spatial configuration of the human body modeled as a graph, with nodes corresponding to anatomical keypoints and edges representing natural connections between body parts. These keypoints typically include joints and salient body landmarks such as shoulders, elbows, hips, and knees. Representing the human body as a graph structure captures its geometric arrangement while abstracting away underlying pixel-level information. \autoref{fig:img_skl_sg_b} illustrates humans poses extracted from \autoref{fig:img_skl_sg_a}.

Such representations have been widely used in computer vision tasks involving human behavior, including activity recognition~\cite{Wang:CVPR:2025}, gesture analysis~\cite{Aich:ICCV:2023}, and interaction modeling~\cite{Li:CVPR:2024}. In the context of CSAI, the relative positioning of individuals, body posture, and inter-subject interactions have been identified as meaningful cues for interpreting potentially abusive situations~\cite{yiallourou2017detection, Kloess:ChallengesCSAM:2019}. Skeleton-based representations thus offer a means of analyzing human pose and interactions without exposing explicit visual content.

To extract human pose information from RCPD, we use the YOLO26~\cite{Sapkota:YOLO26:arXiv} pose model with publicly available weights from the Ultralytics repository. The model is pre-trained on the COCO dataset~\cite{Lin:ECCV:2014} and detects 17 human body keypoints corresponding to major anatomical landmarks. including facial points, upper-body joints, and lower-body joints. These keypoints include nose, eyes, ears, shoulders, elbows, wrists, hips, knees, and ankles. The model predicts the spatial location of each keypoint along with a confidence score indicating the detection reliability.

For each person detected in an image, the model outputs the coordinates of the 17 keypoints in multiple formats, including $2D$ coordinates~$(x, y)$, visibility-aware representations $(x, y, z)$, and normalized coordinate representations. We also store the confidence scores for each detected keypoint. Furthermore, the detector can identify multiple skeleton instances per image when applicable.

These skeleton graphs constitute a second modality of our dataset. Representing individuals using anatomical connections preserves structural information about human pose and interactions while preventing direct access to the original imagery. This enables investigating pose-based patterns and interaction dynamics in CSAI.

\subsubsection{Skeleton Pose Statistics}
\label{sec:skl_statistics}

\autoref{tab:skeleton_summary} summarizes the statistics of the skeleton pose representations in \dataset. The dataset contains a total of 2,135 detected skeleton poses, with a similar distribution across CSAI and non-CSAI subsets. On average, each image contains approximately 1.31 skeletons, indicating that most images include at least one detected person, with occasional scenes containing multiple individuals. Each skeleton contains an average of 11.55 detected keypoints out of the 17 keypoints defined by the pose estimation model. The mean keypoint confidence is 0.669 overall, with slightly lower confidence values observed in the CSAI subset compared to the non-CSAI subset. These statistics indicate that the extracted pose representations capture a substantial portion of the human body structure while maintaining comparable characteristics across both subsets.

\begin{figure}[t]
   \centering
    {\includegraphics[width=0.9\linewidth,clip,trim={0.25cm 0.25cm 0.25cm 0.25cm}]
    {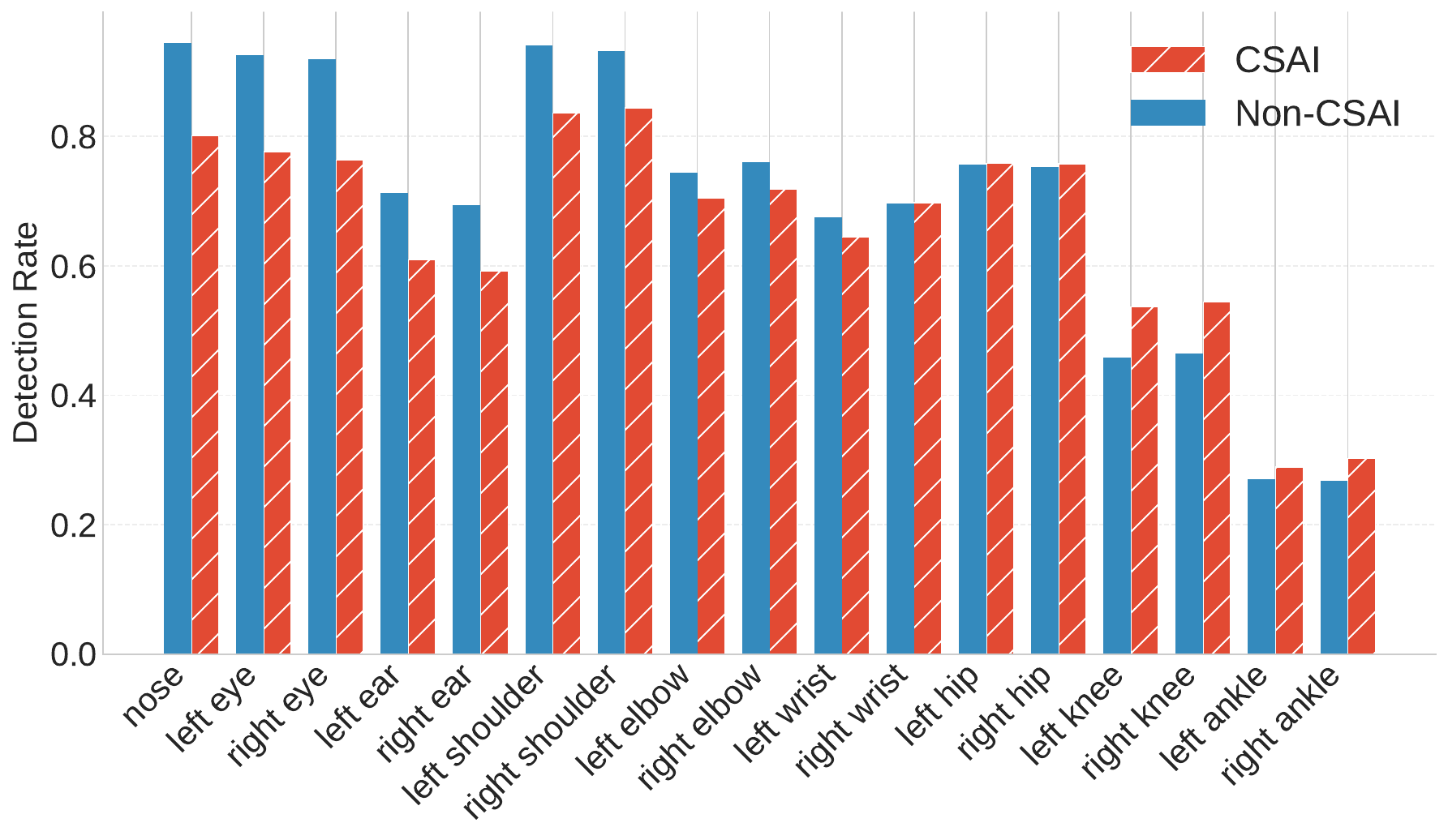}
    }
   \caption{Skeleton pose keypoint detection rate in \dataset.
   }
   \label{fig:histograms_Skl}
   \vspace{-10pt}
\end{figure}

\autoref{fig:histograms_Skl} illustrates the distribution of the keypoint detection rate. Keypoints associated with the face (e.g., \textit{nose}, \textit{eyes}, and \textit{ears}) as well as upper-body joints such as \textit{shoulders} and \textit{hips} exhibit the highest detection rates across both CSAI and non-CSAI images. This behavior is consistent with typical pose estimation patterns, as these joints correspond to larger and more visually distinctive anatomical regions that remain relatively stable across different body poses. In contrast, distal joints such as wrists and ankles tend to be more frequently occluded or outside the image frame, leading to lower detection rates. A similar trend can be observed with higher average confidence scores (additional details are provided in the supplementary material).

\begin{table}[t]
    \footnotesize
    \centering
    \caption{Statistics of the detected skeleton poses in \text{\dataset.}}
    \begin{tabular}{lccc}
        \hline
         & \textbf{All} & \textbf{CSAI} & \textbf{Non-CSAI} \\
        \hline
        Skeleton poses           & 2,135 & 1,065 & 1,070 \\
        Mean skeletons per image      & 1.31 & 1.29 & 1.33 \\
        Mean detected keypoints/skeleton & 11.55 & 11.18 & 11.91 \\
        Mean keypoint confidence      & 0.669 & 0.650 & 0.688 \\
        \hline
    \end{tabular}
    \label{tab:skeleton_summary}
\end{table}

%% file: 5_baseline_experiments.tex
\section{Baseline Experiments}
\label{sec:baseline_experiments}

\subsection{Baseline Method and Experimental Setup}

To assess whether the proposed derived representations are informative for CSAI analysis, we conducted baseline experiments using the ASGRA framework~\cite{Barros:BMVCW:2025}. ASGRA (Attention over Scene Graphs for Sensitive Content Analysis) is an approach designed for CSAI classification that operates on structured graph representations, thus well-suited for evaluating both scene and skeleton graphs.

For experiments using scene graphs, we adopt the feature construction procedure defined in \citet{Barros:BMVCW:2025}. In this approach, node representations are formed by concatenating token embeddings derived from object labels with the normalized bounding-box coordinates of the detected objects. Edge features are obtained from token embeddings of the corresponding predicate labels describing the relationships between object pairs.

Learning and inference are performed using a Graph Attention Network (GATv2~\cite{brody2021attentive}), which dynamically computes attention coefficients for each edge to weight the contributions of neighboring nodes during message passing, thus focusing on the most informative relational triplets for classification. After the message-passing stages, an attention-based pooling layer aggregates node representations into a single graph-level embedding, which is then passed to a multilayer perceptron (MLP) for the final binary classification task (CSAI vs.~Non-CSAI). 

For skeleton-based experiments, the same baseline architecture is employed, adapting the graph input to represent human pose structures. Nodes correspond to body keypoints and edges follow the anatomical connections between joints. Node features are constructed by concatenating three components: token embeddings representing the keypoint type (e.g., nose, shoulder, elbow), the keypoint's spatial coordinates, and the associated confidence score.

To investigate the complementarity between the two modalities, we employ a stacked fusion strategy in which a meta-classifier is trained on heterogeneous signals extracted from both representation models. Specifically, we concatenate three types of information: (i) the graph-level embeddings produced by each GAT encoder, which capture high-level structural representations of the pose and scene graphs; (ii) the positive-class logits generated by each MLP classifier head, which reflect the models’ raw decision scores before normalization; and (iii) the corresponding posterior probabilities obtained via softmax. We then trained an XGBoost~classifier~\cite{Chen:SIGKDD:XGBoost:2016} on these stacked features to learn how to optimally combine the outputs.

We conducted all experiments using a 5-fold cross-validation protocol. Given the sensitivity of the target domain, we report recall as the primary evaluation metric. Recall is critical in this context, as it measures the model's ability to identify CSAI instances correctly and thereby reduces the likelihood of abusive content being overlooked.

\subsection{Quantitative Results}
\label{qualitative_results}

\autoref{tab:validation_results} reports the baseline performance on \dataset\ using different input modalities. The SG-baseline achieves an accuracy of 74.47\% and the highest precision among the evaluated methods, indicating that relational context captured by scene graphs provides strong signals for CSAI classification. This behavior is consistent with the dataset's structural characteristics, in which scene graphs capture contextual cues related to both the presence of children and their interactions with surrounding objects. In particular, object categories associated with minors frequently appear in the graphs (see Section.~\ref{scene_graph_statistics}), while relational predicates encode interactions and scene configurations that may provide additional contextual signals for distinguishing CSAI from non-CSAI scenarios (e.g., the predicate \textit{wearing}).

Regarding the Skl-baseline, it presents lower overall performance, with 64.39\% accuracy. This result should be interpreted in light of both the experimental design and the nature of the data. The skeleton-based model uses the same architecture as scene graphs to maintain methodological consistency across modalities, as our objective is not to design specialized architectures for pose-based representations but rather to evaluate the usefulness of structural signals. Additionally, certain scenarios present inherent limitations for pose-based representations. For instance, some images contain close-up views of body regions, such as genital areas, where reliable full-body pose estimation cannot be obtained, leading to missing or incomplete skeleton detections. Despite these challenges, skeleton-based representations still capture meaningful cues related to human pose and interactions. We illustrated examples of these situations in the supplementary material.

The Fusion-baseline achieves the best overall performance, reaching 75.33\% accuracy and the highest \text{F1-score}. More importantly, it achieves a recall of 74.96\%, improving by approximately 2.5 percentage points over the \text{SG-baseline}. We highlight that recall is particularly critical in CSAI classification, as it measures the model's ability to identify abusive instances correctly and thereby reduces the likelihood of overlooking harmful content. This improvement indicates that scene graph and skeleton representations capture complementary aspects of the visual scene: relational context between objects and pose-related information about human interactions.

\begin{table}[t]
\setlength{\tabcolsep}{2pt}
    \centering 
    \footnotesize
    \caption{Baseline performance on \dataset\ using different input modalities. SG-baseline operates on scene graphs, Skl-baseline uses skeleton, and Fusion-baseline combines both modalities. Results were obtained using 5-fold cross-validation.}
    \begin{tabular}{lccccc}
        \hline
        \textbf{Method}  &  \textbf{Acc. (\%)}  & \textbf{Precis. (\%)} & \textbf{Recall (\%)} & \textbf{F1 (\%)}\\
        \hline
        SG-baseline & $74.47$ \tiny{$\pm~0.01$} & $\textbf{76.96}$ \tiny{$\pm~0.01$} & $72.49$ \tiny{$\pm~0.03$} & $74.31$ \tiny{$\pm~0.02$}\\
        Skl-baseline & $64.39$ \tiny{$\pm~0.01$} & $64.77$ \tiny{$\pm~0.01$} & $67.16$ \tiny{$\pm~0.02$} & $65.66$ \tiny{$\pm~0.01$} \\
        Fusion-baseline  & $\textbf{75.33}$ \tiny{$\pm~0.01$}  & $76.38$ \tiny{$\pm~0.01$} & $\textbf{74.96}$ \tiny{$\pm~0.01$} & $\textbf{75.62}$ \tiny{$\pm~0.01$} \\
        \hline
        \end{tabular}
    \label{tab:validation_results}
\end{table}

\vspace{-3mm}

\paragraph{Prediction disagreement.}To further investigate the complementary behavior between the two modalities, we examine cases where one modality correctly classifies a sample while the other fails. Error analysis is performed by aggregating predictions across test folds within each cross-validation run, ensuring that each sample is evaluated only when it appears in the test partition. This analysis reveals that the two representations capture different aspects of the visual scene. For instance, there are 175 samples where the skeleton-based model correctly predicts the label while the scene graph model fails. Among these, 109 correspond to CSAI instances and 66 to non-CSAI samples. Conversely, there are 350 samples where the scene graph model produces the correct prediction while the skeleton model fails (213 non-CSAI and 137 CSAI). These results indicate that scene graphs and skeleton poses provide complementary signals for CSAI classification.

\vspace{-3mm}

\paragraph{Fusion-baseline error recovery.}We analyze how the fusion model benefits from this complementarity by identifying cases where the fusion approach corrects errors made by the individual modalities. The fusion model correctly classifies 143 samples that are misclassified by the scene graph model (86 CSAI and 57 non-CSAI). Similarly, 372 samples misclassified by the skeleton-based model (162 CSAI and 210 non-CSAI) are correctly classified by fusing both modalities. It amounts to 63.6\% recovery of the skeleton-based errors and 34.9\% of the scene graph errors. These findings provide additional evidence that the two modalities contribute complementary information. In particular, the improvements in recall observed in the fusion results suggest that pose information can help retrieve CSAI cases that are not sufficiently characterized by contextual relationships, while scene-level context helps disambiguate cases where pose information is incomplete or ambiguous.

\subsection{Qualitative Results}

\begin{figure}[t]  
    \centering
    \includegraphics[width=0.85\linewidth]{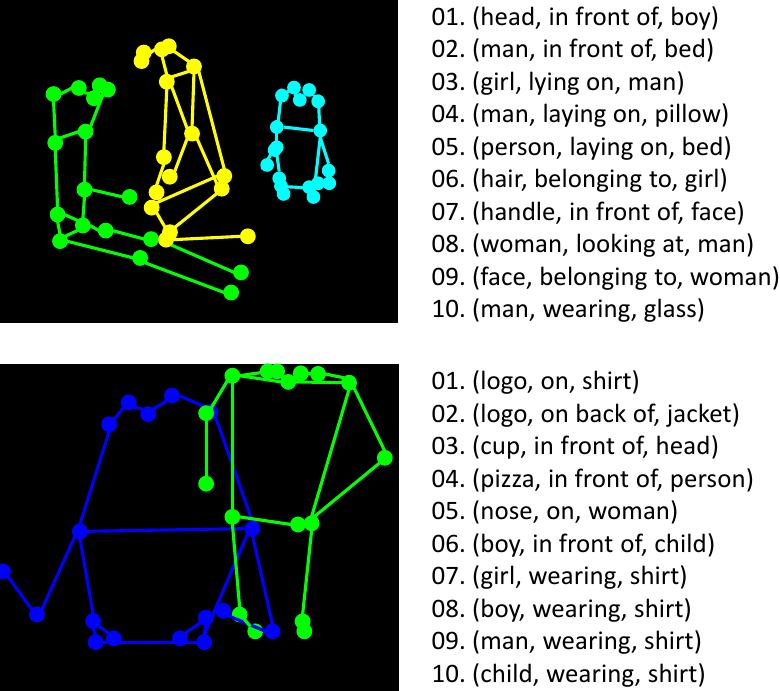}
    \caption{
    Qualitative examples illustrating complementary nature of the representations. Top: a case where pose-based cues help the classification, while scene graphs fail. Bottom: a case where objects and predicates help the classification, while pose fails.
    }
    \label{fig:SG_qualitative}
    \vspace{-5pt}
\end{figure}

\autoref{fig:SG_qualitative} illustrates qualitative examples of the structural representations employed. The top row shows a CSAI sample, where the left image corresponds to the skeleton pose representation and the right side lists the top scene graph triplets extracted from the same image. In the pose representation, three human skeletons are detected, with two of them suggesting a physical interaction where one individual appears to be partially on top of another. The scene graph representation also captures contextual relations consistent with this scenario, including triplets such as \textit{(man, sitting on, bed)}, \textit{(girl, laying on, bed)}, and \textit{(boy, laying on, bed)}, indicating that multiple individuals are positioned on the same surface. Notably, several relations involve entities corresponding to children, such as \textit{girl} and \textit{boy}, which appear directly connected to objects and spatial interactions in the scene. Importantly, relations such as \textit{(man, wearing, glass)} indicates that an accessory is more salient than other pieces of clothing. This type of relation may implicitly suggest a lack of typical clothing coverage, which can be indicative of nudity in the scene.

The bottom row shows a non-CSAI example with its corresponding structural representations. Two skeletons are detected in close proximity, which could potentially appear ambiguous when relying solely on pose information. However, the scene graph triplets provide contextual cues that help characterize the scene as benign. Several relations correspond to common everyday contexts, including \textit{(cup, in front of, head)} and \textit{(pizza, in front of, person)}, indicating the presence of ordinary objects associated with daily activities. In addition, multiple clothing-related predicates appear in the scene graph, such as \textit{(girl, wearing, shirt)}, \textit{(boy, wearing, shirt)}, and \textit{(man, wearing, shirt)}, suggesting that the individuals are fully dressed. 

These examples highlight the complementary nature of the structural representations provided in our \dataset\ dataset. In CSAI, scene graphs may reveal the presence of children and limited clothing, while pose representations emphasize interactions and body configurations. Conversely, in non-CSAI, contextual relations involving everyday objects and clothing items help deem the situation as benign, even when pose configurations appear ambiguous.

%% file: 6_conclusion.tex
\section{Conclusion}
\label{sec:conclusion}

We introduced \dataset, a privacy-preserving dataset composed of structural representations derived from the RCPD dataset maintained by the Brazilian Federal Police. Releasing scene graphs and human pose skeletons rather than the original images enables CSAI classification research respecting the strict legal and ethical constraints.

The representations capture complementary aspects of visual scenes. Scene graphs encode contextual relationships between objects, while skeleton graphs describe human pose and interactions. Experimental results using baseline models showed that both modalities retain useful information for CSAI classification, and that combining them improves performance.

By making these derived representations publicly available, \dataset\ lowers the barrier for researchers to investigate structural and contextual cues associated with CSAI. We hope this dataset encourages further work on privacy-preserving approaches for child safety applications in computer vision and supports the development of more robust methods for classifying harmful content.

%% file: 7_ethical_legal_considerations.tex
\section*{Ethical and Legal Considerations}
\label{sec:ethical_legal_considerations}

Research involving CSAI raises significant ethical and legal challenges. The Brazilian Federal Police maintains the RCPD dataset used in this work, which contains material that is illegal to possess or distribute outside authorized law enforcement environments. Therefore, researchers in this project do not have access to the original images and cannot visualize or store them. 

\vspace{-3.5mm}

\paragraph{Generation of derived representations.} The structural representations released in this work (scene and skeleton graphs) contain only abstract relational and pose information derived from the images, without including raw pixels or any identifiable visual details. Prior to public release, we manually reviewed the generated representations to ensure that they do not contain explicit content or information that could allow reconstruction of the original images.

\vspace{-3.5mm}

\paragraph{\dataset~release.} Our dataset is to be used exclusively for academic research, and access will require providing details on the purpose of the request, the institutional affiliation of the requesting party, as well as agreeing with a use policy developed in partnership with legal and law-enforcement experts on child protection. This documentation will also help monitor the dataset's chain of custody. 

\vspace{-3.5mm}

\paragraph{Potential misuse.}
A potential risk of misuse lies in the application of the released representations and baseline models for harmful purposes. Malicious actors could attempt to leverage these structural representations or models to facilitate the identification or retrieval of CSAI-related patterns from large image collections.

%% file: 9_acknowledgments.tex
\section*{Acknowledgments}
\label{sec:acknowledgments}

This work is partially funded by FAPESP 2023/12086-9, FAEPEX/UNICAMP 2597/23, and the Serrapilheira Institute R-2011-37776. C.~Ernesto (2025/08423-5), A.~Barros (2024/09372-2), and S.~Avila (2023/12865-8, 2020/09838-0, 2013/08293-7) are also funded by FAPESP. C.~Caetano and S.~Avila are also funded by H.IAAC 01245.003479/2024-10. S.~Avila is further supported by CNPq 316489/2023-9.

%% file: 8_supplementary.tex
\clearpage
\setcounter{page}{1}
\maketitlesupplementary

This supplementary material provides additional analyses and examples that complement the results presented in the main paper. In particular, we include qualitative examples of challenges in skeleton pose extraction observed in CSAI scenarios, as well as additional statistics and analyses of the scene graph representations in \dataset. These materials aim to provide further insights into the dataset's structural properties and to illustrate how the proposed representations capture contextual and pose-related information relevant to CSAI analysis.

\label{sec:appendices}

\appendix
\section{Skeleton Pose Extraction}

\begin{figure}[h]
    \centering
    \includegraphics[width=\linewidth]{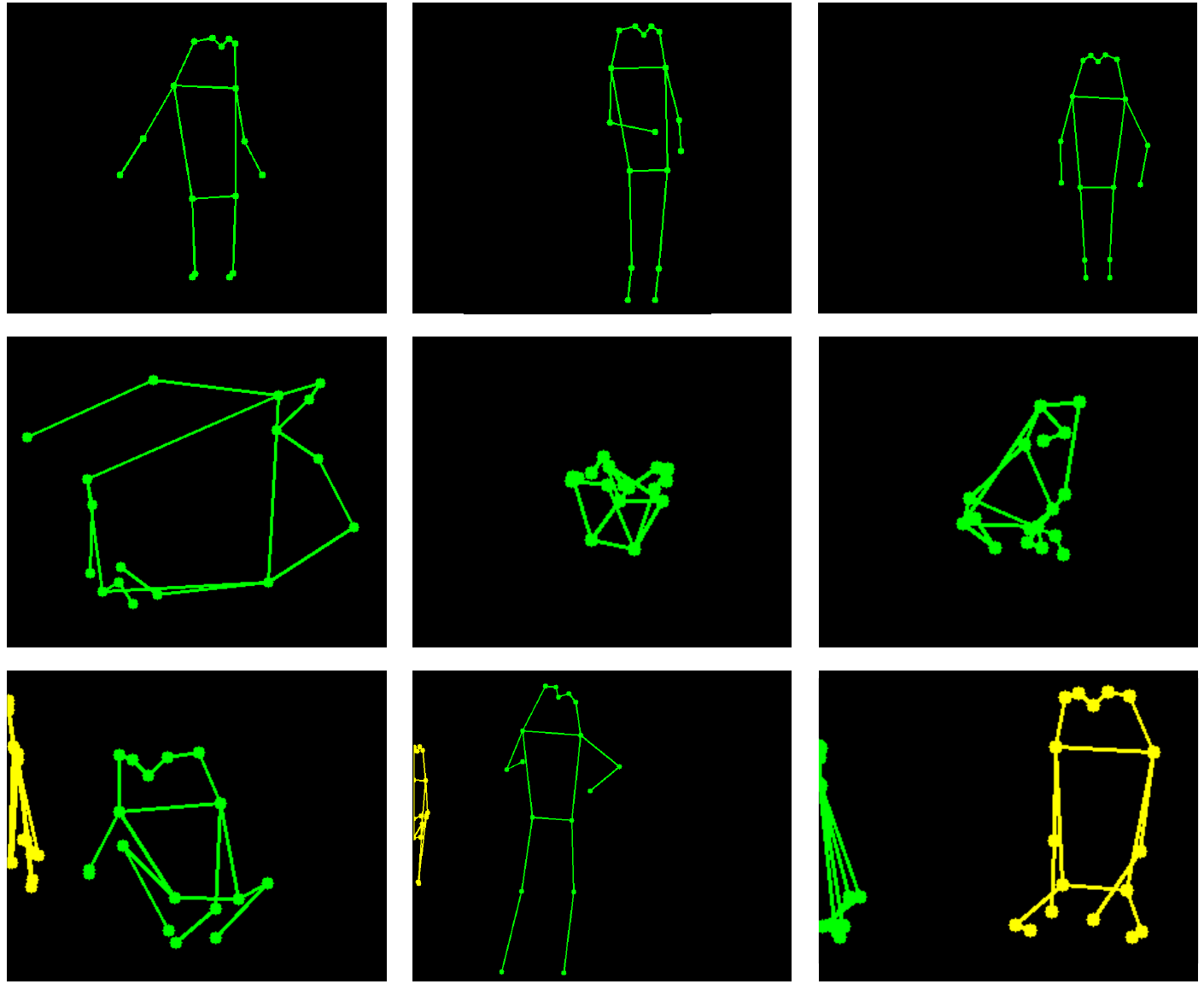}
    \caption{Examples of skeleton pose estimation limitations in CSAI scenarios. The top row shows images with only a single individual, resulting in a single detected skeleton and limited relational pose information. The middle row presents close-up views of specific body regions, where reliable full-body pose estimation cannot be obtained. The bottom row illustrates cases where skeleton extraction is incomplete due to occlusions, truncation, or challenging viewpoints.}
    \label{fig:incomplete_unavailable}
\end{figure}

\paragraph{Challenges.} Pose estimation may present limitations in certain CSAI scenarios. Some images contain only a single individual, which provides limited relational pose information for skeleton-based models. In other cases, the image depicts close-up views of specific body regions, such as genital areas, where reliable full-body pose estimation cannot be obtained. \autoref{fig:incomplete_unavailable} illustrates representative examples of these situations, highlighting cases where skeleton extraction is incomplete or unavailable.

\begin{figure}[h]
   \centering
    \subfloat[Mean confidence score per keypoint]{\includegraphics[width=0.8\linewidth,clip,trim={0.25cm 0.25cm 0.25cm 0.25cm}]
    {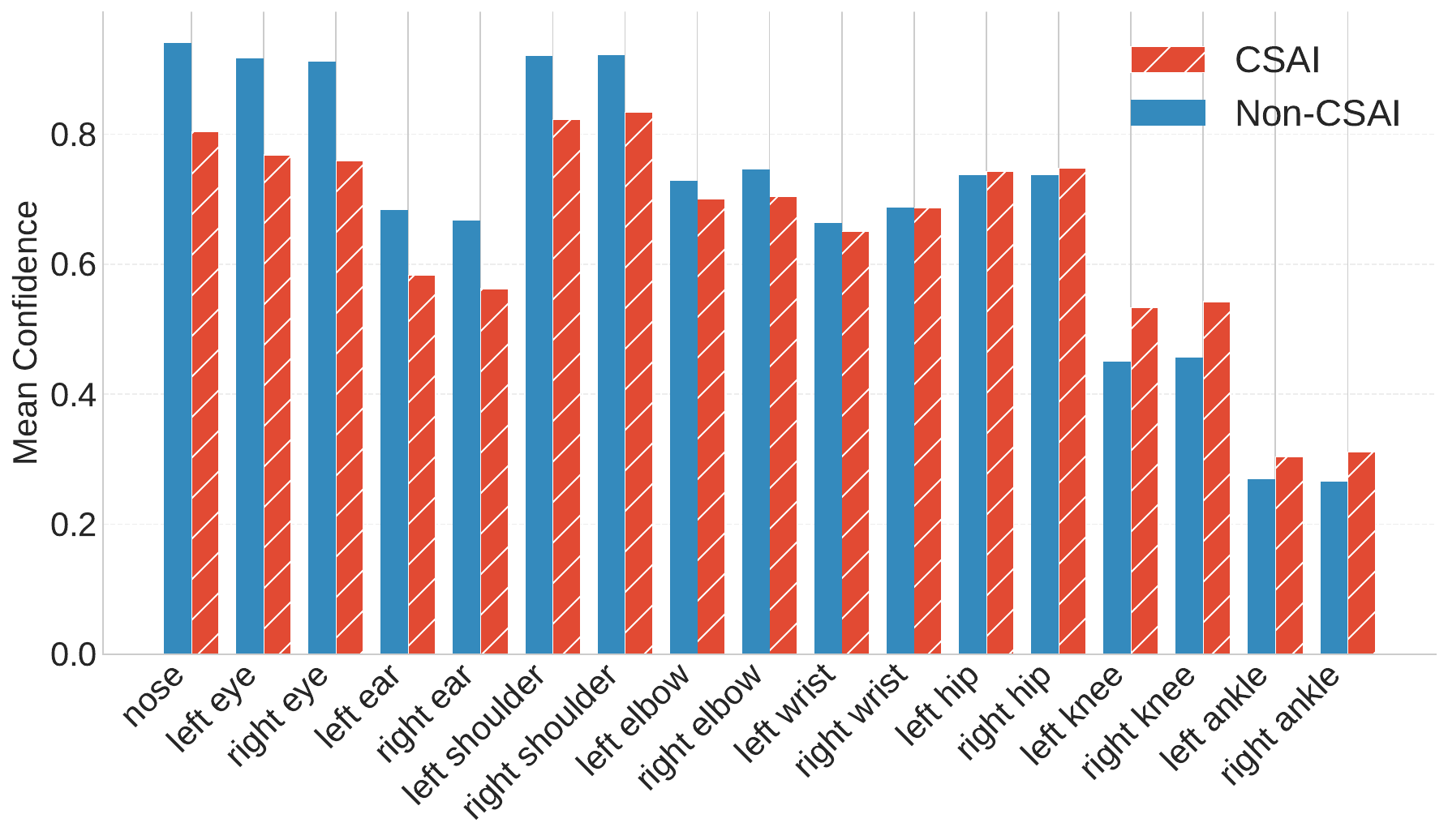}
    }
   \caption{Skeleton pose keypoint statistics in \dataset.
   }
   \label{fig:histograms_Skl_b}
\end{figure}

\paragraph{Average Confidence Scores.} As mentioned in Section~\ref{sec:skl_statistics}, keypoints associated with the face as well as upper-body joints exhibit the highest detection rates across both CSAI and non-CSAI images. O the other side, distal joints such as wrists and ankles tend to be more frequently occluded or outside the image frame, leading to lower detection rates. A similar trend can be observed with higher average confidence scores (\autoref{fig:histograms_Skl_b}). This correlation suggests that the pose estimator is more reliable for central body joints and facial landmarks.

\section{Confusion Matrices of Baseline Models}

\begin{figure*}[h!]
    \centering
    \begin{minipage}{0.3\linewidth}
    \centering
    \includegraphics[width=\linewidth]{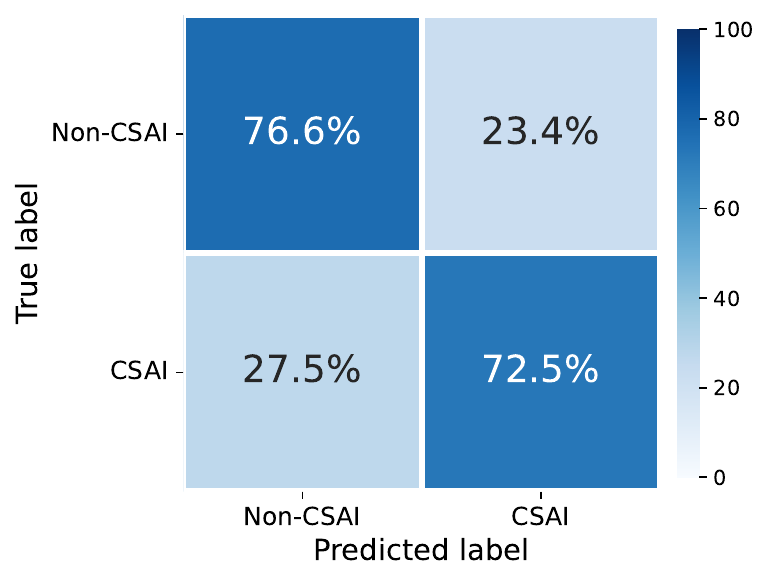}
    (a) SG-baseline (scene graphs)
    \end{minipage}
    \hfill
    \begin{minipage}{0.3\linewidth}
    \centering
    \includegraphics[width=\linewidth]{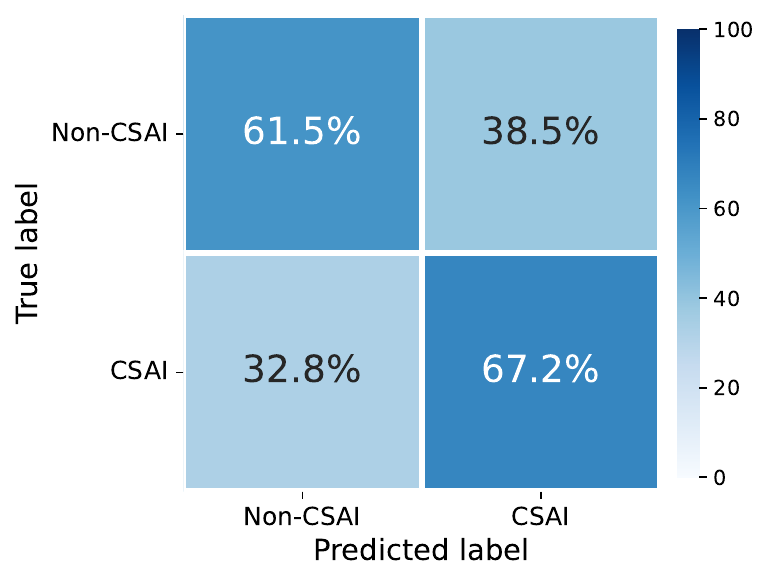}
    (b) Skl-baseline (skeleton poses)
    \end{minipage}
    \hfill
    \begin{minipage}{0.3\linewidth}
    \centering
    \includegraphics[width=\linewidth]{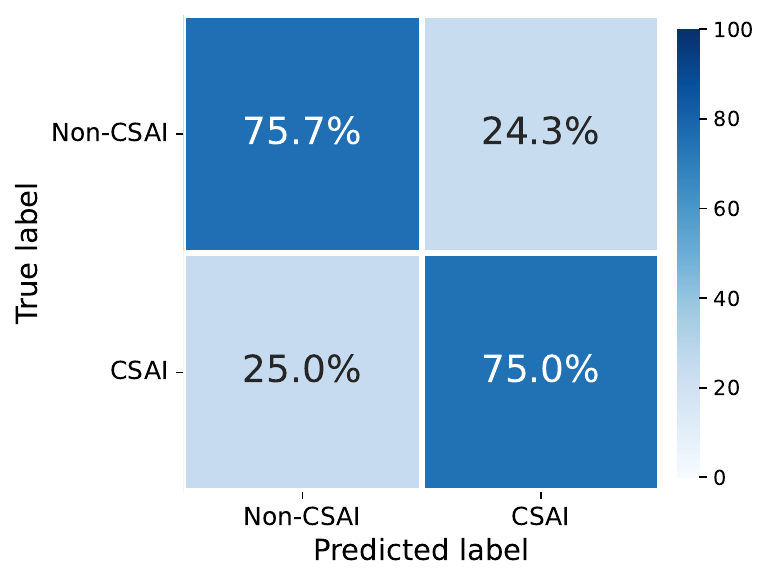}
    (c) Fusion-baseline (both modalities)
    \end{minipage}
    \caption{Confusion matrices of the baseline models evaluated in \dataset.}
    \label{fig:confusion_matrices}
\end{figure*}

\autoref{fig:confusion_matrices} presents the confusion matrices of the baseline models evaluated in Section~\ref{sec:baseline_experiments} of the main paper. The matrices provide a detailed view of classification outcomes for CSAI and Non-CSAI samples, showing the distribution of true positives, true negatives, false positives, and false negatives for each model. The results correspond to those obtained from experiments using 5-fold cross-validation.

As expected from the quantitative results reported in the main paper, the SG-baseline outperforms the skeleton-based model, reflecting the richer contextual information captured by scene graph representations. The Skl-baseline shows a higher number of misclassifications, consistent with the limitations of pose-only representations in certain CSAI scenarios. The fusion model reduces both false negatives and false positives compared to the individual modalities, further supporting the complementary nature of scene graph and skeleton pose representations.

\section{Prediction Disagreement Analysis}

\autoref{tab:rrediction_disagreement} presents a prediction disagreement analysis between the skeleton-based model, the scene graph model, and the fusion model. The results show that the two modalities capture complementary information about the scenes. The fusion model corrects a substantial portion of these errors: 143 cases in which the scene graph model fails and 372~cases in which the pose model fails are successfully recovered by the fusion approach. These results further support the complementary nature of the two structural representations and explain the improved performance observed when both modalities are combined.

\begin{table}[!ht]
    \centering
    \footnotesize
    \caption{Prediction disagreement and correction analysis between skeleton-based, scene graph, and fusion models.}
    \begin{tabular}{lccc}
    \hline
    \textbf{Prediction pattern} & \textbf{Samples} & \textbf{CSAI} & \textbf{Non-CSAI} \\
    \hline
    Pose correct / Scene wrong & 175 & 109 & 66 \\
    Scene correct / Pose wrong & 350 & 137 & 213 \\
    \hline
    Scene wrong / Fusion correct & 143 & 86 & 57 \\
    Pose wrong / Fusion correct & 372 & 162 & 210 \\
    \hline
    \end{tabular}
    \label{tab:rrediction_disagreement}
\end{table}

\paragraph{Error Recovery Metrics.}
In addition to the disagreement analysis described above, we quantify how often the fusion model corrects errors produced by each modality. These measurements correspond to the recovery metrics reported in the main paper (Section~\ref{qualitative_results}) and provide a normalized view of the complementary behavior between the structural representations. Specifically, we define two metrics: the \textit{pose error recovery rate} and the \textit{scene error recovery rate}. These metrics measure the proportion of errors made by each single-modality model that the fusion model subsequently corrects.

Let $E_{\text{pose}}$ denote the total number of samples misclassified by the skeleton-based model, and let $C_{\text{pose}\rightarrow\text{fusion}}$ denote the number of those errors that the fusion model correctly classifies. The pose error recovery rate is defined as:
\begin{equation}
R_{\text{pose}} =
\frac{C_{\text{pose}\rightarrow\text{fusion}}}{E_{\text{pose}}}.
\end{equation}

Similarly, let $E_{\text{scene}}$ denote the total number of samples misclassified by the scene graph model, and let $C_{\text{scene}\rightarrow\text{fusion}}$ denote the number of those errors corrected by the fusion model. The scene error recovery rate is defined as:
\begin{equation}
R_{\text{scene}} =
\frac{C_{\text{scene}\rightarrow\text{fusion}}}{E_{\text{scene}}}.
\end{equation}

These metrics provide a complementary perspective to the disagreement analysis by measuring the relative proportion of errors that can be resolved through multimodal fusion. As discussed in Section~\ref{qualitative_results} of the main paper, the fusion model recovers a substantial fraction of the errors made by the individual modalities, further supporting the complementary nature of the structural representations.

\section{Frequent Scene Graph Triplets in CSAI and Non-CSAI Samples}

\autoref{tab:top_triplets_csai_score} and~\autoref{tab:top_triplets_non_csai_score} present the top-10 most frequent relational triplets observed in the scene graph representations for CSAI and Non-CSAI samples, respectively, ranked by the sum of confidence scores produced by the scene graph generator. Many of the most frequent triplets in both subsets correspond to anatomical relationships, such as \textit{(hair, on, head)}, \textit{(nose, on, face)}, and \textit{(eye, on, face)}, which naturally arise from the frequent detection of human body parts in images containing people. However, differences emerge when considering contextual relationships involving people and surrounding objects. In the CSAI subset, triplets such as \textit{(woman, laying on, bed)}, \textit{(girl, laying on, bed)}, and \textit{(man, laying on, bed)} appear among the most frequent relations, reflecting common spatial configurations present in the scenes. In contrast, the Non-CSAI subset contains several triplets associated with clothing, including \textit{(boy, wearing, shirt)}, \textit{(woman, wearing, shirt)}, and \textit{(girl, wearing, shirt)}, which are absent from the top relations in CSAI samples. This difference is consistent with the observation that predicates related to clothing appear less frequently in CSAI images. Overall, these patterns illustrate how scene graph representations capture structural and contextual cues that may help differentiate CSAI from non-CSAI scenarios while preserving a privacy-preserving abstraction of the original images.

\begin{table}[h]
\footnotesize
    \centering
    \caption{Top triplets in CSAI samples ranked by summed confidence score.}
    \begin{tabular}{lccc}
    \hline
    Triplet & Count & Sum score & Mean score \\
    \hline
    (hair, on, head) & 1270 & 804.478 & 0.633 \\
    (nose, on, face) & 862 & 594.013 & 0.689 \\
    (woman, has, hair) & 685 & 403.372 & 0.589 \\
    (woman, laying on, bed) & 369 & 369.000 & 1.000 \\
    (girl, has, hair) & 627 & 352.883 & 0.563 \\
    (eye, on, face) & 505 & 314.005 & 0.622 \\
    (girl, laying on, bed) & 309 & 309.000 & 1.000 \\
    (ear, on, head) & 400 & 251.421 & 0.629 \\
    (nose, on, head) & 336 & 230.271 & 0.685 \\
    (man, laying on, bed) & 218 & 218.000 & 1.000 \\
    \hline
    \end{tabular}
    \label{tab:top_triplets_csai_score}
\end{table}

\begin{table}[h]
\footnotesize
    \centering
    \caption{Top triplets in Non-CSAI samples ranked by summed confidence score.}
    \begin{tabular}{lccc}
    \hline
    Triplet & Count & Sum score & Mean score \\
    \hline
    (hair, on, head) & 1569 & 1080.801 & 0.689 \\
    (nose, on, face) & 1350 & 1019.411 & 0.755 \\
    (woman, has, hair) & 998 & 563.683 & 0.565 \\
    (eye, on, face) & 732 & 535.964 & 0.732 \\
    (ear, on, head) & 693 & 500.922 & 0.723 \\
    (boy, wearing, shirt) & 826 & 451.690 & 0.547 \\
    (girl, has, hair) & 740 & 424.042 & 0.573 \\
    (woman, wearing, shirt) & 933 & 392.642 & 0.421 \\
    (girl, wearing, shirt) & 830 & 376.818 & 0.454 \\
    (nose, on, head) & 436 & 305.143 & 0.700 \\
    \hline
    \end{tabular}
    \label{tab:top_triplets_non_csai_score}
\end{table}